\documentclass[conference]{IEEEtran}
\usepackage{times}

\usepackage[numbers]{natbib}
\usepackage{multicol}
\usepackage[bookmarks=true]{hyperref}
\usepackage{xcolor}
\usepackage{xspace}
\usepackage{multirow}
\usepackage{listings}
\usepackage{amsmath}
\usepackage{amssymb}
\usepackage{amsthm}
\usepackage{algorithm}
\usepackage{booktabs}
\usepackage{graphicx}
\usepackage{colortbl}
\usepackage{color}
\usepackage{subcaption}
\usepackage{balance}

\usepackage[flushleft]{threeparttable}
\definecolor{citecolor}{HTML}{0071BC}
\hypersetup{colorlinks,linkcolor={red},citecolor={citecolor}}  
\definecolor{lightgray}{gray}{0.93}
\newcommand{\alias}{PhyGrasp\xspace}
\newcommand{\data}{PhyPartNet\xspace}

\usepackage{todonotes}

\newcommand{\eg}{\emph{e.g.}}

\usepackage{listings}
\usepackage{framed}
\begin{document}

\title{\alias: Generalizing Robotic Grasping \\ with Physics-informed Large Multimodal Models}


\author{\authorblockN{Dingkun Guo*, Yuqi Xiang*, Shuqi Zhao, Xinghao Zhu, Masayoshi Tomizuka, Mingyu Ding$\dag$, Wei Zhan
}
\authorblockA{Mechanical Engineering\\
University of California, Berkeley\\ 
{\small \texttt{edu@dkguo.com, xyq87121119@gmail.com, myding@berkeley.edu}}} \\
*These authors contribute equally to this work. \quad \quad $\dag$ Corresponding author and project lead.
}

\maketitle

\begin{abstract}

Robotic grasping is a fundamental aspect of robot functionality, defining how robots interact with objects.
Despite substantial progress, its generalizability to counter-intuitive or long-tailed scenarios, such as objects with uncommon materials or shapes, remains a challenge.
In contrast, humans can easily apply their intuitive physics to grasp skillfully and change grasps efficiently, even for objects they have never seen before.

This work delves into infusing such physical commonsense reasoning into robotic manipulation.
We introduce \alias, a multimodal large model that leverages inputs from two modalities: natural language and 3D point clouds, seamlessly integrated through a bridge module.
The language modality exhibits robust reasoning capabilities concerning the impacts of diverse physical properties on grasping, while the 3D modality comprehends object shapes and parts.
With these two capabilities, \alias is able to accurately assess the physical properties of object parts and determine optimal grasping poses.
Additionally, the model's language comprehension enables human instruction interpretation, generating grasping poses that align with human preferences.
To train \alias, we construct a dataset \data with 195K object instances with varying physical properties and human preferences, alongside their corresponding language descriptions.
Extensive experiments conducted in the simulation and on the real robots demonstrate that \alias achieves state-of-the-art performance, particularly in long-tailed cases, \eg, about 10\% improvement in success rate over GraspNet. 
Project page: \url{https://sites.google.com/view/phygrasp}.

\end{abstract}

\IEEEpeerreviewmaketitle

\section{Introduction}

\begin{figure}[t]
  \centering
   \includegraphics[width=0.99\linewidth]{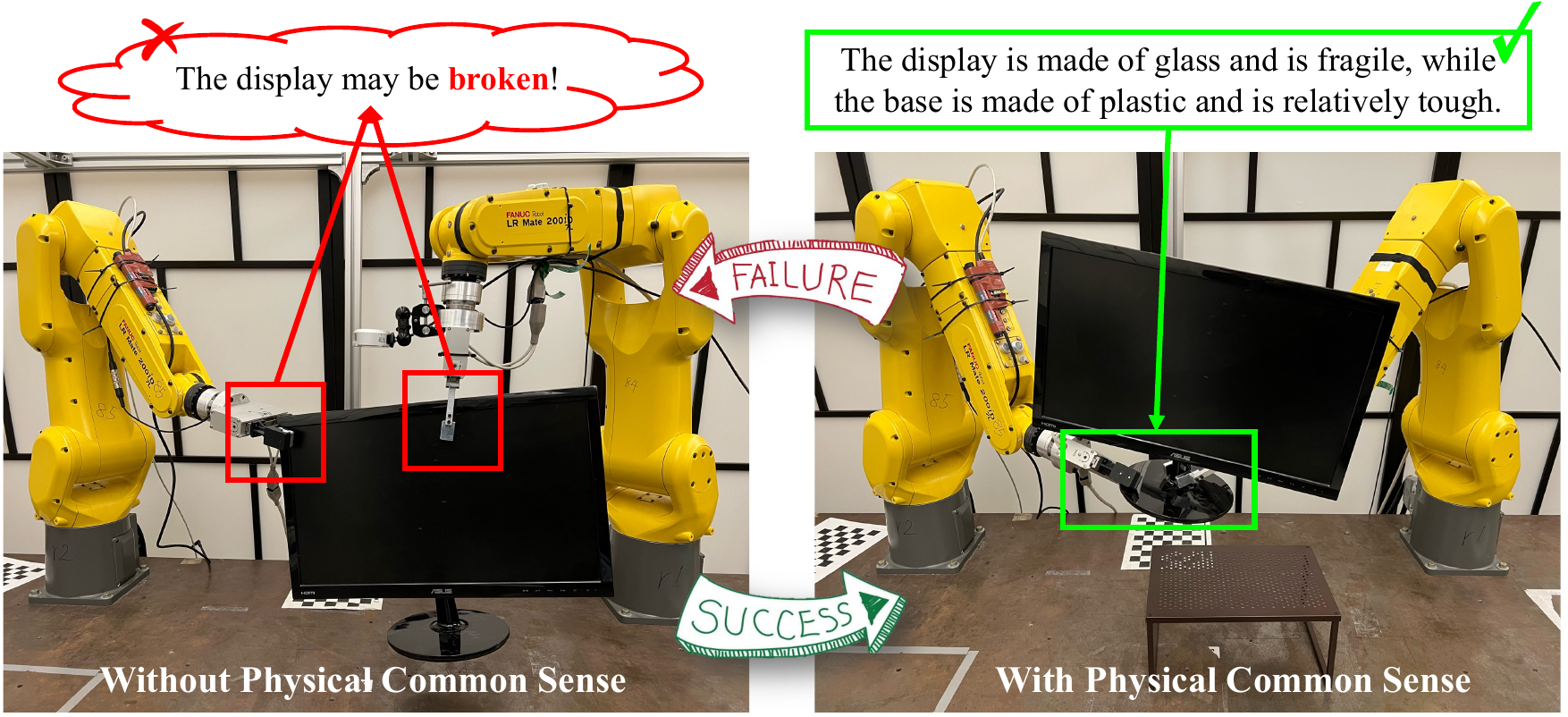}
   \vspace{-16pt}
   \caption{Motivation of \alias. 
Current robot grasping policies (left) typically predict grasping poses based solely on the object's 3D shape, neglecting its physical properties. This oversight can lead to potential damage to the display.
In contrast, integrating physical common sense into robotic systems (right) can address this issue effectively.
   }
   \label{fig:teaser}
   \vspace{-8pt}
\end{figure}

Human-like embodied intelligence represents an important milestone in robot manipulation, offering practical applications such as household robots that can assist with our daily tasks.
Despite notable advancements~\cite{cui2021toward,mason2018toward}, current capabilities of robots still lag far behind humans, particularly in physical commonsense reasoning and generalizability~\cite{billard2019trends}.
Humans possess inherent multimodal reasoning abilities and an intuitive sense of physics, enabling us to accurately plan actions by leveraging commonsense knowledge. It is also intuitive for us to generalize knowledge to uncommon and counterfactual objects or situations.
For example, as illustrated in Fig.~\ref{fig:teaser}, humans intuitively recognize the fragility of the display and understand the need to grasp the base when lifting a monitor, realizing that mishandling it could lead to screen breakage.
Existing robot grasping techniques lacking physical common sense may inadvertently disregard these principles and result in damage. Incorporating physical common sense into robotic systems can mitigate this issue.
Therefore, it becomes an important challenge to empower robots with such capabilities to handle long-tailed objects and scenarios.

Previous methods for robotic grasping and manipulation generally fall into two streams.
1) The first stream directly estimates low-level robot actions or trajectories for execution~\cite{brohan2022rt, rt2, padalkar2023open}. These methods typically rely on large-scale data for training, resulting in models that struggle to generalize to novel scenarios or robot platforms.
2) To improve the generalizability, the second stream~\cite{fang2023anygrasp,fang2023robust} proposes to implement analytical methods or learned models to generate affordance maps or grasping pose proposals. Subsequently, it plans low-level robot actions based on these estimated affordances or poses. The underlying motivation is that grasp poses are easier to generalize than robot action sequences.
Nonetheless, existing grasping pose detection algorithms often focus on the analysis of 3D shapes and semantics of objects, while overlooking part information, physical senses, or constraints. Consequently, they still face challenges in generalizing to objects with diverse physical properties in long-tailed scenarios.
The incorporation of physical commonsense remains a fundamental aspect that is largely unexplored within existing robotic grasping frameworks.

In recent times, the rapid evolution of large language models (LLMs), such as OpenAI's ChatGPT~\cite{chatgpt}, has showcased robust understanding and generalization capabilities, holding promise for physical commonsense reasoning.
However, these models lack perceptual environmental information, such as detailed parts and shapes from 3D vision, which poses a challenge in utilizing LLMs for real grasping applications.
Although some vision-language models (VLMs) have been proposed to provide vision information for LLMs, their focus has predominantly been on visual question-answering tasks, leaving them ill-equipped to reason effectively about the physical world, particularly within domains like robotic grasping and manipulation.
Considering 3D models such as PointNet~\cite{qi2017pointnet} and VoxNet\cite{maturana2015voxnet}, which offer substantial insight into object shapes and poses in the physical world, an intuitive solution emerges: building a multimodal model that bridges the 3D and language modalities. This integration aims to facilitate a comprehensive physical reasoning of objects in robotic grasping tasks.

In practice, it is non-trivial to train an interface between the 3D and language modalities, due to its data-intensive nature and underrepresentation in standard multimodal pretraining datasets.
Existing datasets and benchmarks typically either concentrate solely on grasping without considering the underlying physical concepts (\eg, material, fragility, mass, friction)~\cite{fang2020graspnet}, or they focus on high-level physical understanding without addressing low-level grasping estimation~\cite{gao2023physically}, restricting their usefulness for robotic grasping and manipulation tasks.
Our objective is to address this problem from both sides.

In this work, we construct a physically grounded 3D-language dataset, termed \data. It contains 195K unique object instances featuring various physical properties across their parts based on PartNet~\cite{mo2019partnet}.
For each object instance, we sample physical properties, such as material, fragility, mass, density, and friction, for individual parts of the object. Subsequently, we generate corresponding grasping probability maps using analytical grasping solutions, along with machine-generated language instructions and preferences.

Based on \data, we introduce \alias, a multimodal model designed to serve as an interface between LLMs and 3D encoders, effectively bridging and grounding high-level physical semantics and language into low-level grasping maps.
\alias employs frozen PointNext~\cite{qian2022pointnext} and Llama~2~\cite{touvron2023llama} as its encoders, coupled with a carefully crafted bridge module capable of integrating information from language, visual local, and visual global representations to generate final predictions.
It offers several appealing benefits. 
Firstly, it predicts grasping poses based on both language descriptions and 3D information regarding an object's physical properties, such as material, fragility, mass, density, and friction. 
Secondly, its language comprehension enables the interpretation of human instructions, facilitating the output of grasping poses aligned with human preferences. 
Lastly, it demonstrates strong generalizability to long-tailed, unseen, and even counterfactual objects.

Our primary contribution is \alias, which generalizes robotic grasping through the integration of physics-informed large multimodal models.
For the first time, we facilitate grasping pose detection by leveraging the object's part-level physical properties.
We conduct experiments in the simulation and on real robots to demonstrate the effectiveness of \alias.
Another contribution is our \data dataset, a comprehensive collection of large-scale 3D mesh instances featuring diverse part-level physical properties and corresponding language annotations.
We aspire for our work to inspire future research in robot grasping, particularly among those inclined towards physical reasoning and interactions.

\section{Related Work}

\subsubsection{Physical Reasoning}
Previous work's focus was on estimating the physical properties of objects through visual perception, using interaction data as a primary source of learning~\cite{wu2015galileo, wu2016physics, li2020visual}. A distinct body of research has shifted towards developing representations that encapsulate physical concepts, going beyond direct property estimation~\cite{janner2019reasoning, DensePhysNet,ding2021dynamic,qi2020learning,chen2022comphy,bear2021physion,tung2024physion++}.
Notably, methods~\cite{liu2023minds, li2023can,gao2023physically} explore physical reasoning using LLMs and VLMs, \eg, \cite{gao2023physically} introduces a dataset specifically designed to quantify and enhance object-centric physical reasoning capabilities.
Moreover, OpenScene~\cite{Peng2023OpenScene} employs CLIP~\cite{Radford2021LearningTV} to discern objects within scenes based on properties like material composition and fragility.  
However, they focus on high-level physical understanding without addressing low-level grasping estimation, restricting their usefulness for robotic grasping and manipulation tasks.
This work introduces \data, which not only underpins our methodology but also facilitates advancements in robotic manipulation by providing a more nuanced understanding of physical properties and their implications for robotics grasping.

\subsubsection{Large Multimodal Models}
The community has witnessed the emergence of multimodal large language models (MLLMs), designed to augment the capabilities of traditional language models by incorporating the ability to process and understand visual information~\cite{zhang2023llama-adapter, zhang2023internlmxcomposer, wu2023nextgpt, sun2023emu, alayrac2022flamingo, zhu2023ghost, li2023videochat, lai2023lisa, yang2023gpt-4v, chen2022pali, li2023otter, zhang2023video-llama, ye2023mplugdocowl}. Among these, Flamingo~\cite{alayrac2022flamingo} stands out by utilizing both visual and linguistic inputs to demonstrate impressive few-shot learning capabilities, particularly in visual question-answering tasks.
Building on this foundation, advancements have been made with the introduction of models like GPT-4~\cite{openai2023gpt4}, the LLaVA series~\cite{liu2023llava, lu2023empirical, liu2023improved}, and MiniGPT-4~\cite{zhu2023minigpt4}, enhancing visual language large models (VLLMs) through visual instruction tuning. This innovation has significantly improved these models' ability to follow instructions, a crucial aspect for applications requiring precise interaction with visual content.
Simultaneously, a new wave of models~\cite{wang2023visionllm,peng2023kosmos2,bai2023qwenvl, wang2023allseeing, chen2023shikra} has been developed to strengthen the visual grounding capabilities of VLLMs. These advancements facilitate more nuanced tasks such as detailed region description and precise localization, underscoring the growing sophistication of these systems in interpreting and interacting with visual data.
Despite these significant strides in the development of MLLMs and their enhanced ability to integrate and interpret multimodal data, there remains a notable gap in their application to physical reasoning, particularly in the context of robotic grasping. This gap highlights a pivotal area for future research, where the potential for MLLMs to contribute to the understanding and execution of complex physical interactions can be further explored and realized.

\subsubsection{Large Models for Robot Learning}
Leveraging large pre-trained models holds promise for creating capable robot agents. Numerous works focus on using language models for planning and reasoning in robotics~\cite{huang2022language,saycan,chen2023open,progprompt2023,huang2023visual,raman2022planning,song2023llm,liu2023llm+,vemprala2023chatgpt,ding2023task,palme,yuan2023plan4mc,hu2023tree,lu2023multimodal,wang2023voyager,radford2019language,zhou2023generalizable}. To enable language models to perceive physical environments, common approaches include providing textual descriptions of scenes~\cite{huang2022inner,zeng2022socratic,progprompt2023} or access to perception APIs~\cite{codeaspolicies2022}. Vision can also be incorporated by decoding with visual context~\cite{huang2023grounded} or using multi-modal language models that directly take visual input\cite{palme,openai2023gpt4,embodiedgpt,yang2023octopus}. In this work, we leverage the generalizability of vision and language models for physical common sense reasoning, thereby for the first time, enabling physics-informed robotic grasping.

\subsubsection{Grasp Pose Detection}
The domain of vision-guided grasp pose detection has become a focal point in robotics research, representing a shift from traditional top-down grasping techniques to the thorough exploration and implementation of six degrees of freedom (6 DOF) grasping methods. This evolution is underscored by notable contributions in the field, exemplified by advancements documented in~\cite{mahler2017dex, zhu2022learn, mahler2019learning} for planar grasping. It is further propelled by the introduction of sophisticated 6 DOF methodologies in studies such as~\cite{breyer2021volumetric, jiang2021synergies, zhu20216}. Central to this progression is the development of state-of-the-art 6 DOF grasp pose detection models, particularly exemplified by AnyGrasp~\cite{fang2023anygrasp}. AnyGrasp extracts and encodes geometric features of objects from point clouds, achieving a success rate in object grasping that parallels human capabilities. Leveraging the grasp poses identified by AnyGrasp, subsequent research endeavors have been proposed, concentrating on specific object grasping~\cite{liu2024ok, ju2024robo}. These efforts have extended to articulated object manipulation tasks~\cite{yu2023gamma} as well. However, these investigations often assume fixed physical parameters of objects or aim to identify a universally robust grasp amidst varying physical uncertainties. Such assumptions may lead to impractical or hazardous grasping scenarios, particularly when dealing with delicate object parts, a challenge exacerbated by the limited ability of vision sensors to discern material properties. To address these challenges, an innovative approach is proposed: integrating physical parameters into the grasp planning algorithm through natural language descriptions from human guidance. This approach allows the network to adjust its planning outcomes based on the articulated physical characteristics of objects, thereby enhancing the practicality and safety of robotic grasping operations.

\section{Dataset Generation}

\begin{figure*}[t]
  \centering
   \includegraphics[width=0.99\linewidth]{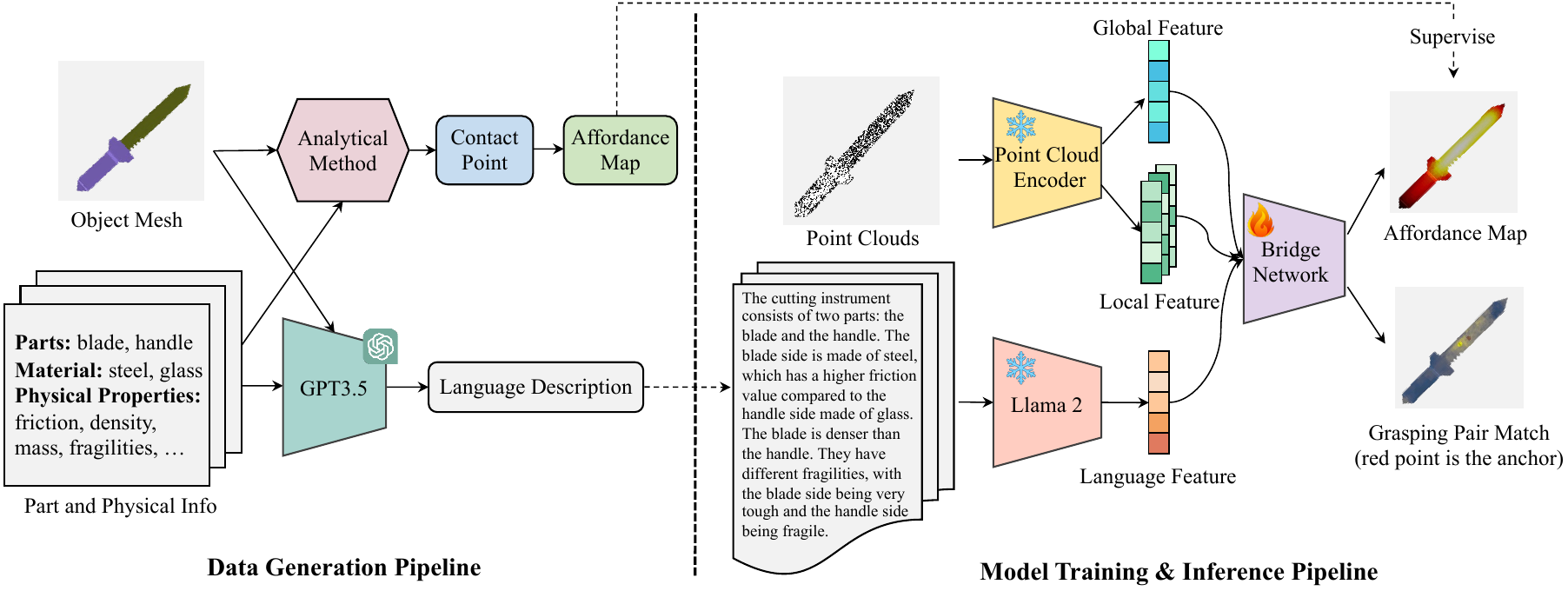}
   \caption{An overview of our \data generation pipeline and our \alias framework. Given object meshes sampled from PartNet, we leverage GPT-3.5 and an analytical method to automatically generate the grasping affordance map and language descriptions for the object instance.
   The generated data is then human-verified, forming our \data.
   We freeze PointNext~\cite{qian2022pointnext} and Llama 2~\cite{touvron2023llama2} and tune the bridge network during training on \data. After training, \alias is able to generalize to novel 3D point clouds and new natural language instructions.
   }
   \label{fig:framework}
\end{figure*}

We develop a dataset, \data, that enables robots to learn physical reasoning for grasping objects. This dataset includes object point clouds, language summaries, and corresponding analytical grasping solutions. The left part of Fig.~\ref{fig:framework} summarizes the data generation process. For each object, we generate multiple instances where different parts of the object have different physical properties (e.g., material, density, mass, friction). Sec.~\ref{sec:data-stats} provides details of the statistics of the dataset. We use analytical methods (refer to Sec.~\ref{sec:analytical}) to calculate force closure grasp pairs and construct a grasping affordance map, which serve as the ground truth grasping solution. As described in Sec.~\ref{sec:gpt}, we use OpenAI's GPT-3.5 to provide descriptive summaries of objects, highlighting different physical properties in each grasp instance.

\begin{figure}[t]
  \centering
   \includegraphics[width=0.495\linewidth]{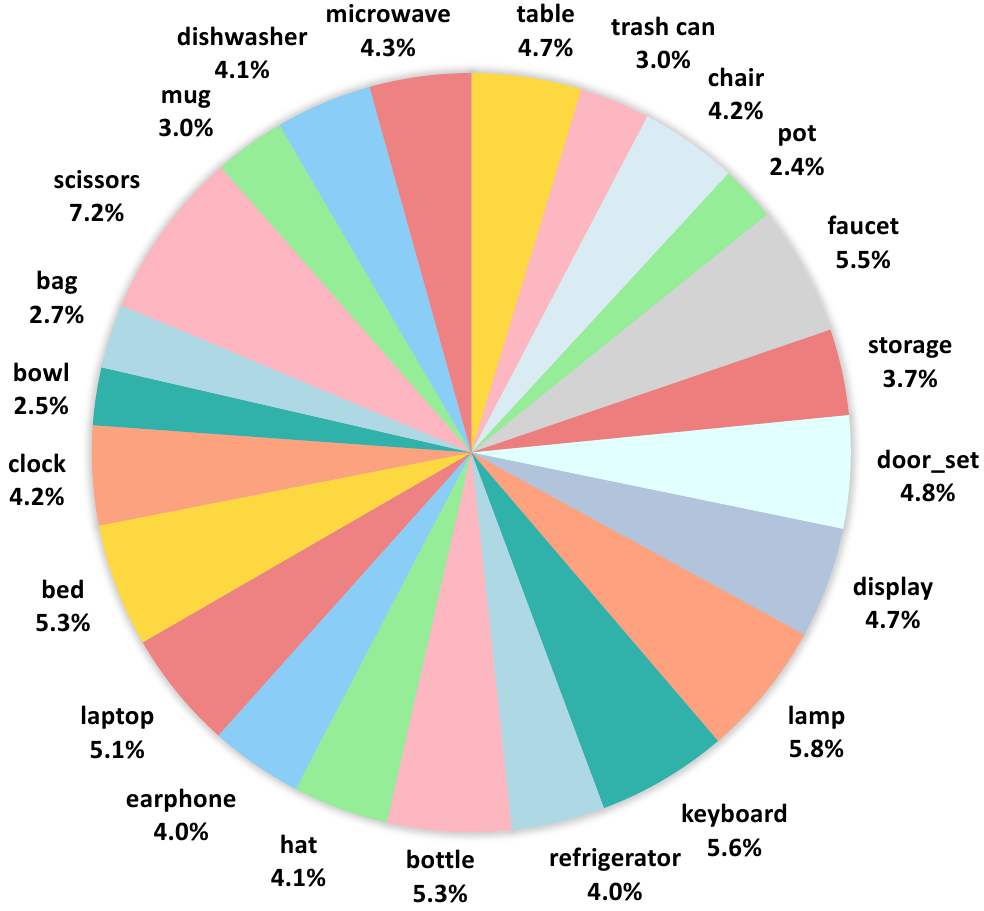}~
   \includegraphics[width=0.495\linewidth]{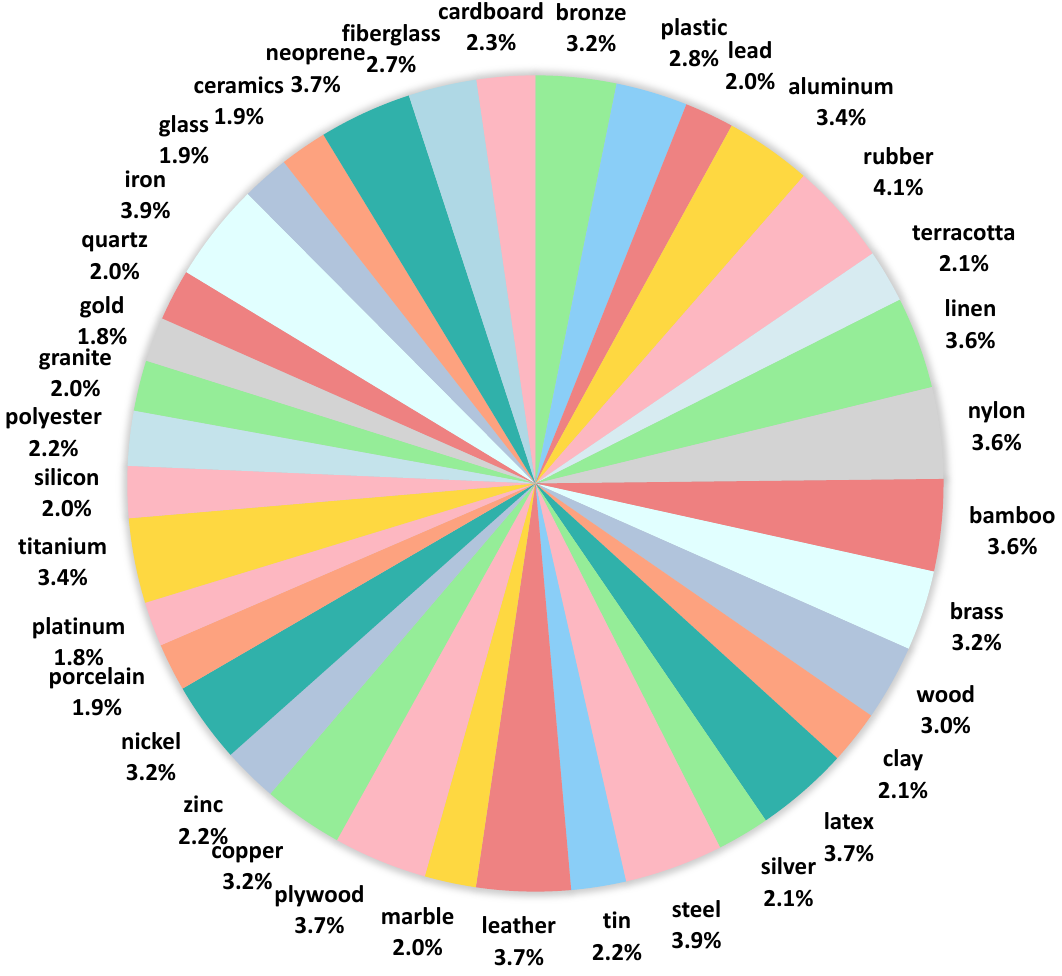}
   \caption{Dataset Statistics. The left and right figures denote instance distributions among objects and materials, respectively.
   }
   \label{fig:object}
\end{figure}

\subsection{Dataset Statistics}
\label{sec:data-stats}
We build our dataset based on the PartNet dataset~\cite{mo2019partnet}, which comprises 28,599 objects across 24 categories, each featuring part segmentation. For every object, we generate multiple instances, varying the materials of different parts. We introduce 16 materials, each associated with unique physics properties: density, friction, and fragility. These properties enable us to compute the mass, center of mass, and maximum normal force applicable to each surface of the object. Additionally, we assign varying levels of grasping probability to each part, reflecting human common sense (\eg~human will not grasp knife blade). In total, we create 193,856 unique instances, with equal distribution among objects and materials (refer to Fig.~\ref{fig:object}).

The training, validation, and testing set have 173,856, 10,000, and 10,000 instances, respectively. In addition, we prompt GPT (refer to Listing~\ref{lst: hardset_prompt}) to pick a ``hard set'' that is a subset of the testing set and contains 370 the most counter-intuitive instances. 

\subsection{Analytical Grasping Solutions}
\label{sec:analytical}
A grasp, denoted by $g$, achieves force-closure if, for any external wrenches (i.e., forces and torques, $F_{ext}$) applied to the object, there exist contact forces $f_c$ within the contact friction cone $K_g$ that counterbalance the external wrenches, satisfying $G f_c = F_{ext}$. Here, $G$ represents the grasp mapping matrix, which is contingent upon the grasp's location, $g$, and the magnitude of $f_c$ can be arbitrarily large~\cite{roa2015grasp}.

In this study, we identify potential grasp candidates by employing a ray-shooting technique around the object to determine contact pairs, thereby conceptualizing a parallel grasp, $g$. The grasp mapping matrix $G$ is then formulated based on $g$'s positioning relative to the object's center of mass (CoM). To assess the force-closure property of a grasp, we employ the optimization problem in Eq.~\ref{eq: force_opt}.

\begin{equation}
\label{eq: force_opt}
\begin{aligned}
\min_{f_c} & \left || f_c \right || \\
s.t. & \  G f_c = F_{ext} \\
& \ f_c \in K_g
\end{aligned}
\end{equation}

While we can ascertain the force-closure status using simpler methods, as indicated in~\cite{mahler2017dex, zhu20216}, the formulation in Eq.~\ref{eq: force_opt} enables the incorporation of additional constraints reflective of the object's physical characteristics. Specifically, we consider the maximum permissible contact force ($\left | f_c \right | \leq \epsilon$), the variation in friction coefficient across the object's surface ($K_g \propto \mu$), and adjustments to the object's center of mass ($G \propto \textrm{CoM}$).

We compute the feasibility of the solution to Eq.~\ref{eq: force_opt} to verify whether a grasp pair is force-closure and complies with other physical prerequisites.

With analytical grasp pairs, we create a grasp affordance map by allocating a Gaussian distribution to each grasping location. The normalized sum of each point on the object mesh represents the grasping probability and follows a mixture of Gaussian distributions. The left column of Fig.~\ref{fig:affordance} illustrates the resulting affordance map. We also save 2,048 points sampled from the surface of each object mesh as object point cloud for each instance for future vision processing.

\begin{lstlisting}[
float=t,
language=bash,
floatplacement=htbp,
frame=TRBL,
frameround=tftf,
belowskip=-2\baselineskip,
basicstyle=\ttfamily\scriptsize,
breakatwhitespace=true,
breaklines=true,
captionpos=b,
columns=flexible,
keepspaces=true,
tabsize=2,
showspaces=false,
showstringspaces=false,
showtabs=false,
label={lst: gpt_prompt},
caption= An example of the prompt for GPT to generate the language summary for an object instance,
abovecaptionskip=0pt,
belowcaptionskip=0pt]
<@\textbf{Role:} you are a grasping analytical assistant, skilled @>
<@in summarizing the features of different objects and @>
materials with a natural language. 
<@~~You should provide as much information as possible with@>
minimal words. 
<@~~You focus on the important features of every part with @>
their materials, rather than the specific values. 
<@~~You will be given a paragraph describing the object and @>
its parts with their materials, densities, frictions, 
fragilities, and human grasp probabilities hint. 
<@~~\textbf{You should follow such rules:}@>
1. Names: Describe the object & material names precisely.
2. Densities: Point out the densest part or the lightest 
part. If the density difference is not obvious, you can 
ignore it.
3. Frictions: Point out the part with the highest 
friction and the lowest friction. If the friction 
difference is not obvious, you can ignore it.
...
<@~~\textbf{I will give you some examples.}@>
<@~~\textcolor{teal}{examples ...}@>
<@\textbf{Instruction:} Please process the following paragraph.@>
<@Output in one paragraph.@>
\end{lstlisting}

\begin{lstlisting}[
float=t,
language=bash,
floatplacement=htbp,
frame=TRBL,
frameround=tftf,
belowskip=-2\baselineskip,
basicstyle=\ttfamily\scriptsize,
breakatwhitespace=true,
breaklines=true,
captionpos=b,
columns=flexible,
keepspaces=true,
tabsize=2,
showspaces=false,
showstringspaces=false,
showtabs=false,
label={lst: language_example},
caption= A human example of the language description for GPT prompt.,
abovecaptionskip=0pt,
belowcaptionskip=0pt]
<@\textcolor{teal}{Input:}  There is a faucet, it has several parts including a @>
<@    switch, a frame, and a spout. The material of each part @>
<@    is plastic, brass, and fiberglass, with friction: 0.4, @>
<@    0.38, 0.6, density: 1400, 8530, 2020, fragility: normal, @>
<@    tough, normal.@>

<@\textcolor{teal}{Output:} The faucet has three parts: switch, frame, and @>
<@    spout. The spout is made of fiberglass with the highest @>
<@    friction. The switch's material is plastic and the frame @>
<@    is made of brass with the highest density.@>
\end{lstlisting}

\begin{lstlisting}[
float=t,
language=bash,
floatplacement=htbp,
frame=TRBL,
frameround=tftf,
belowskip=-1\baselineskip,
basicstyle=\ttfamily\scriptsize,
breakatwhitespace=true,
breaklines=true,
captionpos=b,
columns=flexible,
keepspaces=true,
tabsize=2,
showspaces=false,
showstringspaces=false,
showtabs=false,
label={lst: hardset_prompt},
caption= Prompt with GPT for hard set selection. Object instances in category A is the hard set.,
abovecaptionskip=0pt,
belowcaptionskip=0pt]
<@\textbf{Role:} You should judge the following description@>
<@based on the given information. The description is about@>
<@a specific object and its parts. You should judge@>
<@whether this object is category A or catergory B.@>
<@~~\textbf{Here are some characteristics of A and B:} @>
<@~~1. Material: A (uncommon) vs. B (common); @>
<@~~2. Friction: A (higher) vs. B (lower); @>
<@~~3. Density: A (heavier) vs. B (lighter); @>
<@~~4. Fragility: A (fragile) vs. B (durable); @>
<@~~5. Grasp Guidance: A (specific) vs. B (general). @>
<@\textbf{Instruction:} Please process the following paragraph.@>
<@Answer 'A' or 'B'. If you are not sure,@>
<@ please answer 'I don't know'.@>
\end{lstlisting}

\subsection{Language Summary Generation}
\label{sec:gpt}
For objects composed of multiple parts with varying materials and physicial properties, we utilize OpenAI's GPT-3.5 to generate language descriptions that summarize each instance, emphasizing the distinct physical properties in every grasp scenario. Listing~\ref{lst: gpt_prompt} and Listing~\ref{lst: language_example} illustrate the prompts and examples we provide for GPT to help it understand the relevant terms. GPT generates language summaries for the remaining object instances in our dataset.

\section{Learning Methods}
With our dataset, we are able to train a neural network for robotics grasping grounding with physical reasoning. The training starts with a large vision model and a large language model (refer to Sec.~\ref{sec:feature-extraction}), both of which work in tandem to encode our dataset into visual and linguistic features. We then construct a bridge network (refer to Sec.~\ref{sec:bridge}) that takes these features as input and yields a grasping affordance map, as well as a complementary classifier to generate an array of corresponding grasping pairs for each point on an object’s point cloud. Sec.~\ref{sec:loss} details the losses we use for training.

\subsection{Feature Extraction}
\label{sec:feature-extraction}
\subsubsection{Vision Encoder}
We use the PointNeXt architecture~\cite{qian2022pointnext} to transform an object's point cloud into global and local visual features. With a PointNeXt encoder pre-trained on the ModelNet40 dataset~\cite{Wu_2015_CVPR}, we extract a global feature vector with a shape of (1024,) for each object's point cloud. Since ModelNet40 dataset contains different objects from those in our dataset, these global features facilitate our model's ability to generalize to objects out of the domain of our dataset. For the extraction of local features, we leverage the PointNeXt encoder in conjunction with its part segmentation decoder, outputting local features of dimension (64,) for each point in an object's point cloud. The encoder-decoder pair, having been trained on ShapeNet—the same dataset that underpins our work—embeds detailed part segmentation information within the local features, enhancing our network’s capacity to discern the variations among different parts of an object.

\subsubsection{Language Encoder}
We utilize Llama~\cite{touvron2023llama} to encode the language descriptions of each instance into linguistic features. Opting for the representation from the model's 20th layer, as indicated by the findings in~\cite{zou2023interfacing}, which demonstrated optimal outcomes for feature extraction, we obtain features with a dimension of (4096,).

\subsection{Bridge Network}
\label{sec:bridge}

\begin{figure}[t]
  \centering
   \includegraphics[width=0.99\linewidth]{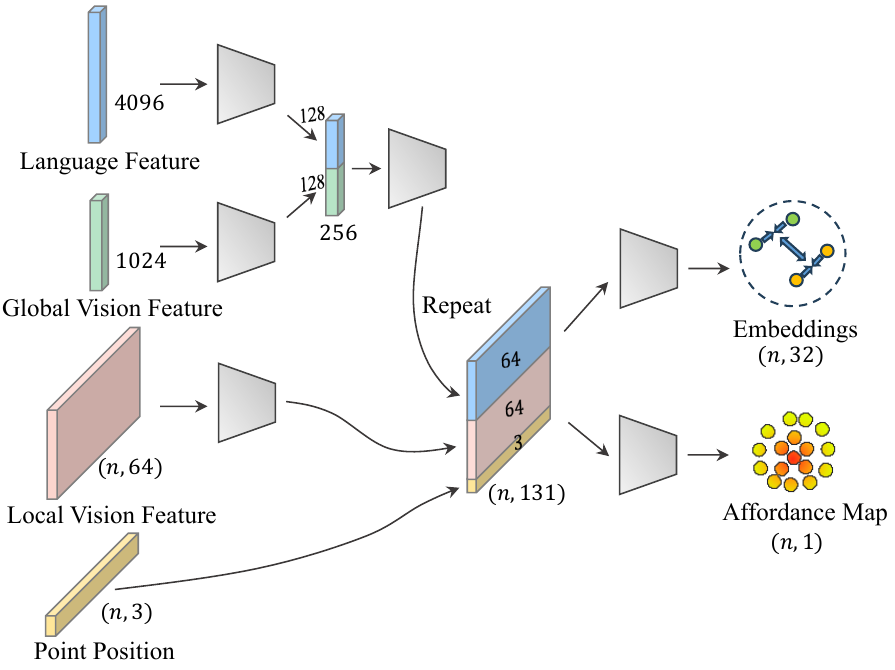}
   \vspace{-8pt}
   \caption{The architecture for the bridge module of \alias. It outputs the grasping probability (affordance map) and the pair embedding for each point.
   }
   \label{fig:bridge}
\end{figure}

Our bridge network uses the extracted features to predict grasping solutions. Fig.\,\ref{fig:bridge} illustrates the structure of our bridge network. We use a multilayer perceptron (MLP) to compress both the global visual and linguistic features down to a dimension of (128,) and mix them with another MLP to generate a global feature of (64,). In a parallel process, we refine the local visual feature through an MLP and amalgamate the output with the object's global features and point cloud, culminating in a composite feature vector of (64+64+3,) for each point on the point cloud. We then deploy another two distinct MLPs: one functions as a predictor to generate a grasp affordance map, and the other acts as a classifier to identify corresponding grasp pairs using embeddings.

\subsection{Losses} 
\label{sec:loss}
We use three different loss functions for training. The first is a global loss, $L_{\text{g}}$, computing the difference between predicted affordance map and the groud truth:

\begin{equation}
\label{eq: lg}
\begin{aligned}
L_{\text{g}} = \frac{1}{N} \sum_{i = 1}^{N} ||G_i - G_i^{\text{gt}}|| 
\end{aligned}
\end{equation}
where $N$ is the number of instances, $G_i$ is the $i$th output affordance map by our model, and $G_i^{\text{gt}}$ is the corresponding ground truth in our dataset.

The second loss function, $L_{\text{emb}}$, is for the construction of embeddings to distinguish positive and negative grasping pairs. It is a linear combination of its positive part $L_{\text{emb}}^{\text{p}}$ and the negative part $L_{\text{emb}}^{\text{n}}$ as following:
\begin{gather}
\label{eq: lemb}
L_{\text{emb}} = \lambda L_{\text{emb}}^{\text{p}} + L_{\text{emb}}^{\text{n}} \\
L_{\text{emb}}^{\text{p}} = \frac{1}{N} \sum_{i = 1}^{N} \frac{1}{K_i^{\text{p}}} \sum_{k = 1}^{K_i^{\text{p}}} [|| Q_{i, k, 1} - Q_{i, k, 2} || - \delta_p]_{+}^2 \\
L_{\text{emb}}^{\text{n}} = \frac{1}{N} \sum_{i = 1}^{N} \frac{1}{K_i^{\text{n}}} \sum_{k = 1}^{K_i^{\text{n}}} [\delta_n - || Q_{i, k, 1} - Q_{i, k, 2} ||]_{+}^2 
\end{gather}
where $K_i^{\text{p}}$ and $K_i^{\text{n}}$ are the numbers of positive and negative grasp pairs for $i$th instance, respectively. $Q_{i, k}$ is our model's embeddings output for the $k$th grasp pair of $i$th instance. $\delta_{p}$ and $\delta_{q}$ are respectively the margins for the positive and negative embedding loss. $[x]_+$ = $\max(0, x)$ denotes the hinge.

The third loss, $L_{\text{match}}$, is for finding grasping pair match. As mentioned in Sec.~\ref{sec:bridge}, we use a MLP to predict each point pair's matching probability with embeddings. We define this probability as $M_{i, k}$ for the $k$th grasping pair in the $i$th instance. Correspondingly, $M_{i, k}^{\text{gt}}$ is matching score in the ground truth, which equals $1$ for positive pairs and $0$ for negatives:

\begin{equation}
\label{eq: lc}
\begin{aligned}
L_{\text{match}} = -\sum_{i = 1}^{N} \sum_{k = 1}^{K_i^{\text{p}} + K_i^{\text{n}}} \log{p(M_{i, k} = M_{i, k}^{\text{gt}})} 
\end{aligned}
\end{equation}
where $\log{p(\cdot)}$ is the cross-entropy loss that measures the difference between the predicted matching score and the ground truth.

In the training process, we balance these three loss functions with Automatic Weighted Loss (AWL) \cite{kendall2018multi} to allow combined training of the grasping affordance predictor and the grasping pair classifier.

\section{Experiments and Results}

We conduct experiments in both simulation (Sec.~\ref{sec:sim}) and the real world (Sec.~\ref{sec:real}) to evaluate the performance of our method in robotic grasping.

\begin{figure}[t]
  \centering
   \includegraphics[width=0.99\linewidth]{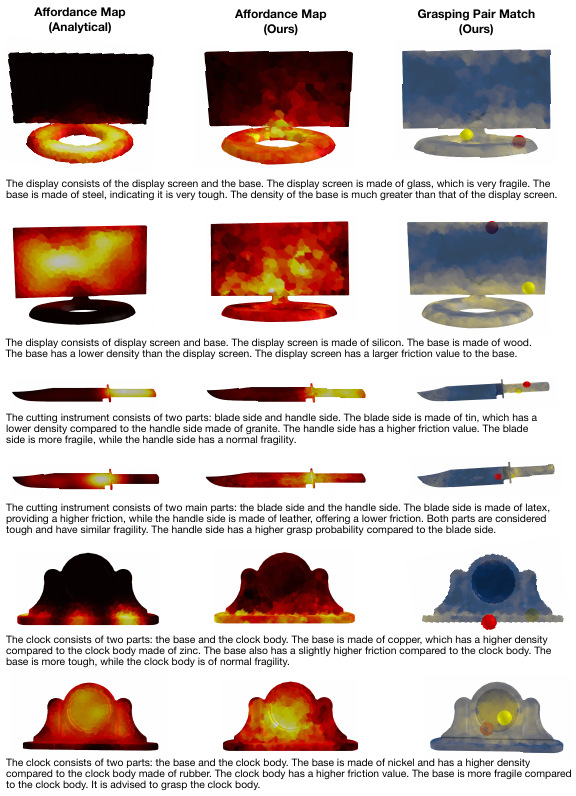}
   \caption{Visualizations of the affordance map and grasping pair match map for our method.
   The left column is the affordance map of the analytical method (ground truth), the middle is our affordance map, and the right is the grasping pair match map. We observe that our affordance map prediction exhibits high quality and closely resembles the ground truth.
   In the match map, yellow intensity indicates the matching confidence, with red and yellow points representing an anchor and its top-1 matching pair.
   }
   \label{fig:affordance}
\end{figure}

\subsection{Simulation Experiments}
\label{sec:sim}

\subsubsection{Settings}
We conduct simulated experiments using PyBullet~\cite{coumans2019}. We implement two gripper fingers to pinch an object at the predicted grasping positions, directed towards each other. For each object instance, we evaluated different models' top-$n$ predictions, considering any among~$n$~trial where the object remained secure between the fingers as a successful grasp.

\subsubsection{Baselines}
\begin{itemize}
    \item \textbf{Analytical (upper bound)} refers to the analytical grasping solutions in each object instance. Evaluating this baseline helps in quantifying the gap between analytical predictions and their practical simulation outcomes.
    \item \textbf{GraspNet}~\cite{fang2020graspnet} is a baseline for general object grasping. It uses a convolutional neural network to predict grasp instances directly from point clouds, providing a comprehensive and efficient approach to robotic grasping.
    \item \textbf{Volumetric Grasping Network (VGN)}~\cite{breyer2021volumetric} constructs a Truncated Signed Distance Function (TSDF) representation of the scene and outputs a volume of the same spatial resolution, similar to the grasping affordance map.
\end{itemize}

\subsubsection{Results}
Table~\ref{table:performace} summarizes the grasping success rate evaluated in simulation for the baseline models and ours. Our model outperforms all baselines in every metric. VGN underperforms on our dataset, particularly with large objects with multiple surfaces like tables, chairs, and beds, due to the difficulty in constructing TSDFs for these items. Additionally, its heavy reliance on visual features makes it prone to failure in scenarios where physical factors alter grasping strategies. GraspNet exhibits slightly lower performance than ours in the general test set, and its performance drops by more than 5\% on the hard set, whereas our model maintains its effectiveness. Since the hard set includes the most counter-intuitive examples, this indicates that our model effectively comprehends language descriptions, while reasoning about physical properties to adapt grasping strategies. In contrast, GraspNet, which depends solely on vision, is likely to struggle with these long-tailed edge cases. 

\begin{table}[t]
\caption{The grasping success rate (\%) evaluated in the simulation for baseline models and our model.
}
\centering
\small
\setlength{\tabcolsep}{8pt}
\begin{threeparttable}
\begin{tabular}{lcccc}
\toprule
\multirow{2}{*}{ Method }   & \multicolumn{2}{c}{ General Set } & \multicolumn{2}{c}{ Hard Set } \\ 
   & Top-1 & Top-5 & Top-1 & Top-5  \\  
\cmidrule(lr){1-1} \cmidrule(lr){2-3} \cmidrule(lr){4-5}
\cellcolor{lightgray}Analytical (upper bound) &  \cellcolor{lightgray}78.0 &  \cellcolor{lightgray}92.1   & \cellcolor{lightgray}70.0 & \cellcolor{lightgray}87.6      \\

GraspNet~\cite{fang2020graspnet} & 56.4  &	83.2 & 50.5 & 77.6      \\
VGN~\cite{breyer2021volumetric} & 34.1 & 45.9 & 33.6 & 43.5     \\
\alias (Ours)  &  \cellcolor{white} \textbf{61.5} \cellcolor{white} &	\cellcolor{white}\textbf{86.0}   & \cellcolor{white}\textbf{59.7} & \cellcolor{white}\textbf{79.2}     \\
\bottomrule     
\end{tabular}
\end{threeparttable}
\label{table:performace}
\end{table}

We also provide qualitative results for our model's predictions of afforance map and grasping pair match in Fig.~\ref{fig:affordance}. Visually, these predictions closely resemble analytical solutions. We explore the impact of physical properties on the same object. For example, the top clock features high friction and low fragility at its base, while the base of the bottom clock is low in friction and is fragile. Our model successfully captures this information and identifies the correct part to grasp. The grasping pair match highlights the efficiency of our embedding and classifier, with the anchor and query points forming a force-closure grasp, enhancing the grasping success rate.

\begin{table}[t]
\caption{Comparisons of grasping affordance map accuracy under different metrics.}
\centering
\small
\setlength{\tabcolsep}{12pt}
\begin{threeparttable}
\begin{tabular}{lcccc}
\toprule

Method &	KLD $\downarrow$ &	SIM $\uparrow$ & AUC-J $\uparrow$\\
\midrule
VGN	& 5.2622	& 0.4452	& 0.5026 \\
Ours &	0.3783 &	0.7306 &	0.8545\\

\bottomrule     
\end{tabular}

\end{threeparttable}
\label{table:affordance}
\end{table}

In Table~\ref{table:affordance}, we report the Kullback-Leibler Divergence (KLD) \cite{fang2018demo2vec}, the Similarity metric (SIM) and the Area Under the Curve (AUC-J) \cite{bylinskii2018different, judd2009learning} to evaluate the effectiveness of the predictions of affordance map. These metrics compare predicted affordance map with the ground truth. The results indicate that our method outperforms VGN in generating more accurate grasping affordances.

\subsubsection{Ablations}

\begin{table}[t]
\caption{Ablation study of our model. We report the grasping success rate (\%) evaluated in the simulation.}
\centering
\small
\setlength{\tabcolsep}{10pt}
\begin{threeparttable}
\begin{tabular}{lcccc}
\toprule
\multirow{2}{*}{ Method }   & \multicolumn{2}{c}{ General Set } & \multicolumn{2}{c}{ Hard Set } \\ 
   & Top-1 & Top-5 & Top-1 & Top-5  \\  
\cmidrule(lr){1-1} \cmidrule(lr){2-3} \cmidrule(lr){4-5}
\cellcolor{white}Ours  &  \cellcolor{white} \textbf{61.5} \cellcolor{white} &	\cellcolor{white}86.0   & \cellcolor{white}\textbf{59.7} & \cellcolor{white}79.2     \\
Ours w/o Local & 46.4 & 81.3 & 44.6 & 76.5 \\
Ours w/o Global & 60.2 & 86.3 & 53.5 & \textbf{79.4} \\
Ours w/o Language & 61.0 & \textbf{86.7} & 55.1 & 77.8 \\

\bottomrule     
\end{tabular}

\end{threeparttable}
\label{table:ablation}
\end{table}

\begin{table}[t]
\caption{The top-5 grasping success rate (\%) evaluated in the real world for GraspNet and our model.}
\small
\setlength{\tabcolsep}{5pt}
\centering
\begin{threeparttable}
\begin{tabular}{lccccc}
\toprule
  Method       &  Scenario & Banana &	Hammer &	Bottle & Overall\\
\cmidrule(lr){1-1} \cmidrule(lr){2-2} \cmidrule(lr){3-5} \cmidrule(lr){6-6}
\multirow{2}{*}{GraspNet} & Normal & 0.2 &	0.2 &	1.0 & 0.5	\\
        & Challenging & 0.0 & 0.2 & 0.0 & 0.2 \\
\multirow{2}{*}{Ours} & Normal & 0.4 & 0.6 & 1.0 & 0.7\\
     & Challenging & 0.6 & 0.6 & 1.0 & 0.7\\

\bottomrule     
\end{tabular}

\end{threeparttable}
\label{table:realexp}
\end{table}

\begin{figure*}
    \centering
    \includegraphics[width=0.99\linewidth]{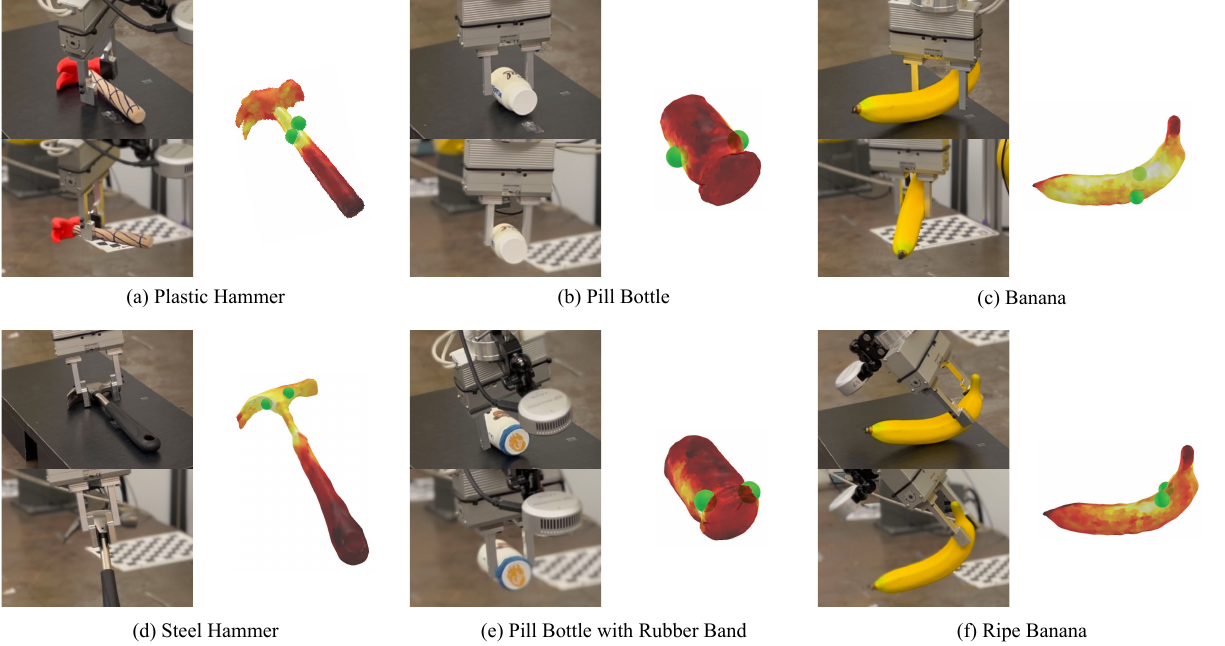}
    \caption{
    Real-world experiments. We select three representative objects with different physical properties.
    \alias accurately predicts locations that align with our expectations for these test scenarios. 1) It effectively estimates the center of gravity of a hammer with various materials and plans graspings. 2) It recognizes that grasping the rubber portion of the pill bottle provides greater stability.
    3) Different language prompts lead to different grasping predictions (high fragility for the body of a ripe banana) even with the same 3D shape, demonstrating the effectiveness of PhyGrasp and its natural human interaction capacity.}
    \vspace{-6pt}
    \label{fig:realworld}
\end{figure*}

\begin{figure}[t]
    \centering    
    \includegraphics[width=0.95\linewidth]{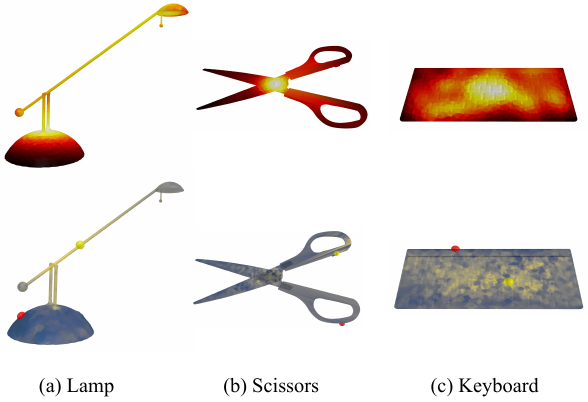}
    \caption{Examples of main sources of failure cases. These failures illustrate long-tailed challenges in robotic grasping, pointing towards solutions like enhanced language descriptions, improved part segmentation, and a broader object variety.
    }
    \label{fig:failure}
    \vspace{-6pt}
\end{figure}

Table~\ref{table:ablation} presents the ablation study of our model. We remove each of important features to evaluate the grasping success rate in the simulation.
\begin{itemize}
    \item Ours w/o Local: Eliminating local vision features significantly impacts our model's capability to discern part segmentation. It hinders the model from prioritizing grasping the parts with a higher probability of successful grasp, leading to the most notable performance drop.
    \item Ours w/o Global: Excluding global features results in a relatively minor impact on the performance. This is understandable since the encoder is pretrained on ModelNet40, which differs from our objects. While this approach aids in generalizing to unseen objects, as demonstrated in our real-world experiments, it does not explicitly cause problems in simulation tests.
    \item Ours w/o Language: Omitting language features leads to the least performance changes in the testing set but results in large failures in the hard set. In most instances, the model can rely on vision features to identify safe grasps. However, in counter-intuitive instances, language information becomes crucial to ensure successful grasping.
\end{itemize}

\subsection{Real-world Experiments}
\label{sec:real}

\subsubsection{Settings}
We conduct experiments with three objects, each offering standard and challenging grasping scenarios (see Fig.~\ref{fig:realworld}). For bananas, the standard test involves unrestricted grasping, while the challenge is to grasp the stem of an overripe banana without damage. The pill bottle challenge requires grasping the cap, bound by a rubber band, avoiding its low-friction body. Hammers presents varying CoM challenges: the robot grasps the uniformly distributed hammer at the center, but has to grip the mass-heavy steel head hammer by its head due to gripper wrench limits.

We use Reality Composer on an iPhone 13 Pro to create objects' meshes and sample point clouds from meshes as input to GraspNet and our model. In normal scenarios, our model receive object name as simple language descriptions, while in the challenging situations, we provide with detailed descriptions outlining our specific grasping requirements. The experiments operate under the assumption of known accurate object pose, as object pose estimation is not the focus of this study. We use PyBullet for motion planning and command a FANUC Robot LR Mate 200iD/7L to grasp object at the predicted grasp positions. 

\subsubsection{Results}
We test top-5 grasping success rates for five trails in each scenario. Our model consistently surpasses GraspNet. Table~\ref{table:realexp} presents a summary of these success rates. Our model achieves a success rate of 70\% in both normal and challenging scenarios, whereas GraspNet attains 50\% in normal conditions and 20\% in challenging ones. This highlights our method's efficacy and dependability in real-world grasp generation.

Fig.~\ref{fig:realworld} illustrates the resulting grasping poses and affordance map prediction. The successful grasping of bananas and hammers further demonstrates our model's ability to generalize to objects that are unseen in training data.

\subsection{Failure Cases and Future Work}
Figure~\ref{fig:failure} presents three failure examples, each representing one of the three main failure case categories.

\begin{itemize}
\setlength\itemsep{0.5em}
    \item Challenges with Uncommon Shapes: The model has difficulty predicting grasp points on uniquely shaped objects. Even with decent affordance map predictions, the embedding classifier fails due to the difficulty in grasping a thin, long-necked lamp.
    \item Overgeneralized Part Segmentation: Our dataset, derived from PartNet, suffers from oversimplified segmentation. For example, scissors are split into only two parts--handle and screw--hindering the model's grasp adjustment based on specific physical properties. 
    \item Grasping Pair Mismatch: The embedding classifier sometimes mismatches two points on the same surface as a graspable pair. Improving the loss function to account for point positions during training and adding a post-processing step to remove same-surface predictions could enhance performance.
\end{itemize}

These failure cases highlight some long-tailed issues the model aims to address. Possible solutions include more specific language descriptions, enhanced part segmentation, and increasing the variety of object types in the dataset. 

\section{Conclusion} \label{sec:conclusion}

This work delves into the integration of physical commonsense reasoning into robotic grasping.
We introduce \alias, a large multimodal model that combines inputs from two modalities: natural language and 3D point clouds, seamlessly connected through a bridge module.
The language modality demonstrates robust reasoning capabilities regarding the impacts of diverse physical properties on grasping, while the 3D modality comprehends object shapes and parts.
By leveraging these two capabilities, \alias accurately evaluates the physical properties of object parts and determines optimal grasping poses. Moreover, its language understanding ability allows for the adjustment of grasping based on human instructions and common sense preferences.
To train \alias, we curate our \data dataset comprising 195,000 object instances with varying physical properties, along with corresponding language descriptions of these properties and human preferences.
We anticipate that our dataset and models will prove to be valuable resources for the community, particularly for those interested in advancing physical reasoning and grasping.

\balance
\bibliographystyle{plainnat}
\bibliography{references}

\end{document}